# ENTERPRISE DOMAIN ONTOLOGY LEARNING FROM WEB-BASED CORPUS

Andrei Vasilateanu, Nicolae Goga, Elena-Alice Tanase, Iuliana Marin
Department of Engineering in Foreign Languages
University Politehnica Bucuresti
Romania
andrei.vasilateanu@upb.ro, n.goga@rug.nl, alice.tanase@gmail.com, marin_iulliana@yahoo.com

*Abstract*— Enterprise knowledge is a key asset in the competing and fast-changing corporate landscape. The ability to learn, store and distribute implicit and explicit knowledge can be the difference between success and failure. While enterprise knowledge management is a well-defined research domain, current implementations lack orientation towards small and medium enterprise. We propose a semantic search engine for relevant documents in an enterprise, based on automatic generated domain ontologies. In this paper we focus on the component for ontology learning and population.

## I. INTRODUCTION

In the current competing and fast-changing corporate landscape, modern enterprises need to maximize the usage of all of their assets. The capacity for adapting and innovating is dependent on the accumulated and transmitted know-how. As more and more employees become "knowledge workers" knowledge is more distributed and is generated in a greater quantity, faster. The high turnover employee rate means that a large amount of this precious knowledge is lost if not captured at enterprise level.

Enterprise knowledge management studies how knowledge can be made reusable and accessible to the enterprise as a whole by formally managing knowledge resources, usually using advanced information technology. [1]

Knowledge can be explicit or implicit, also called tacit knowledge, which represents in fact the vast majority of the knowledge body. Tacit knowledge is distributed, unstructured and can be found in resources such as emails, letters, workflows and even in the interaction between employees, verbally or using collaborative tools. In the Knowledge Creating Company [2] Nonaka proposes a methodology for unlocking the creative forces hidden in the tacit knowledge by a spiral process of socialization, externalization, combination and internalization.

A compounding factor for the knowledge managing problem faced by enterprises is also the explosion of data. Not only is relevant knowledge hidden in unstructured documents, but the quantity of stored documents is also beyond the capacity of a knowledge worker to process manually. Especially a new employee will find it very difficult to search a document repository, which lacks the context required for each document.

Novel techniques for knowledge representation, sharing filtering and discovery are needed. [3] Many of the available solutions, such as using data warehouses, are prohibitive for small and medium enterprises, considered the main economic growth engine.

Our paper presents the domain ontology learning component for a semantic search engine for documents in small and medium enterprises. In Section 2 we present the state-of-the-art for search engines using ontologies and ontology learning. In Section 3 we present the draft architecture of our project and in Section 4 our results in experimenting with ontology learning. The paper ends with the drawn conclusions.

## II. STATE-OF-THE-ART

### A. Semantic search engines

Search engines have untangled The Web, indexing the increasing number of web pages in order to respond almost instantaneously to user queries. However the vision of the Semantic Web [4] in which intelligent agents could also understand data, context and deliver relevant results, is not yet a reality.

The purpose of a search engine is to increase precision and recall where *precision=(number of retrieved relevant documents)/(number of retrieved documents)* and *recall=(number of retrieved relevant documents)/(number of relevant documents)*. Semantic search engines try to maximize these two attributes by exploiting the semantic context of query terms with the relevant domain knowledge. Domain knowledge is formalized in ontologies. Ontology, in computer science, is the formal definition of types, properties and relations between entities in a domain. [5]

Different relations can be expressed in ontologies such as: synonymy, if two concepts denote the same resource, homonymy, if a concept denotes at least two resources, hypernymy and hyponymy between general and specific concepts, meronymy and holonymy between parts and aggregates.

Ontologies can be represented in different languages, the most used being OWL and RDF. Data stored in such formats, using the underlying relationships, can be searched using graph search, with more expressive power than relational search.

Reference [6] describes different methods and examples for implementing semantic search. The main advantage of using a semantic search backed by domain ontology is that the initial query can be automatically rewritten to a) include related terms (concepts which have a relation to concepts in the initial query); b) trim terms in the query that limit the search results, and c) substitute original terms with related terms. The graph representation of ontology terms can also be used to find the strength of the relation between two terms, for example by computing the graph distance between them. Another advantage is using the context of the search for refining the search results, for example favoring the concepts related to the user profile.

Several search engines were built using these techniques such as: SHOE [7] Inquirus [8], TAP [9].

*B. Ontology learning*

The problem with using a domain ontology to guide searches is that domain ontologies are hard to develop manually and it is even harder to find off-the-shelf ontologies [10]. While there exist indexes of existing ontologies and even search engines for ontologies [11], the results of searching are not reliable. It is practically impossible for an arbitrary domain to find a ready-to-use ontology for industrial applications.

The other approach, and the one that we will use in our semantic search engine, is to automatically generate domain ontologies. As we wish to use our enterprise search engine in any domain, we cannot guarantee that we can find structured data sources, perhaps existing ontologies, so we will refer to the most generic approach, generating ontologies from unstructured text. We must point out that generating domain ontologies from text can be very useful not only for semantic search engines, but also for any task involving topic classification such as the one presented in [12].

In [13] there are described the main tasks in the ontology learning. Using Natural Language Processing (NLP) and Machine Learning (ML) algorithms concepts, relations (sub-classing and other) and individuals are discovered from a corpus constructed from unstructured documents. There are few software platforms implementing all of these techniques and the ones that exist are more oriented towards research than industrial use. We have chosen to use for our semantic search engine the platform Text2Onto [14], which is the most aligned with the methods presented in [13]. It also has the advantage of implementing a Possible Ontologies Model (POM), a representation-agnostic way of storing intermediate results, with their associated relevance probability.

### III. SEMANTIC SEARCH ENGINE ARCHITECTURE

The semantic search engine presented in this paper can be used in any small and medium enterprise, as the domain ontology can be changed according to the domain. The search engine is used by knowledge workers to find relevant documents from the company`s document repository. The relevance is computed by comparing the concepts from the query with the concepts in the document, using a domain ontology. The domain ontology itself is generated automatically from public sources on the Web. This solution is amenable for small and medium enterprises as for large enterprises more complex and expensive solutions exist such as data warehouses, which are not affordable by SMEs.

Our high-level architecture is presented in Fig. 1. We detail three different use cases: searching for a document, building/refining the ontology and indexing the central document repository of the company. The last two are continuous processes.

For building the initial ontology and refining it, the Text Crawler component retrieves unstructured text documents from public sources, based on an initial seed term such as "insurance", performs a basic pre-processing and adds them to the Unstructured Text Corpus which is a simple text repository. Our modified Text2Onto component scans the corpus for change and iteratively builds the domain ontology and stores it both as a probabilistic model for further refinement and as RDF/OWL triples in the RDF database for indexing. Although the process is automatic, a domain expert has the possibility to make amends to the ontology by using the Ontological Editor component.

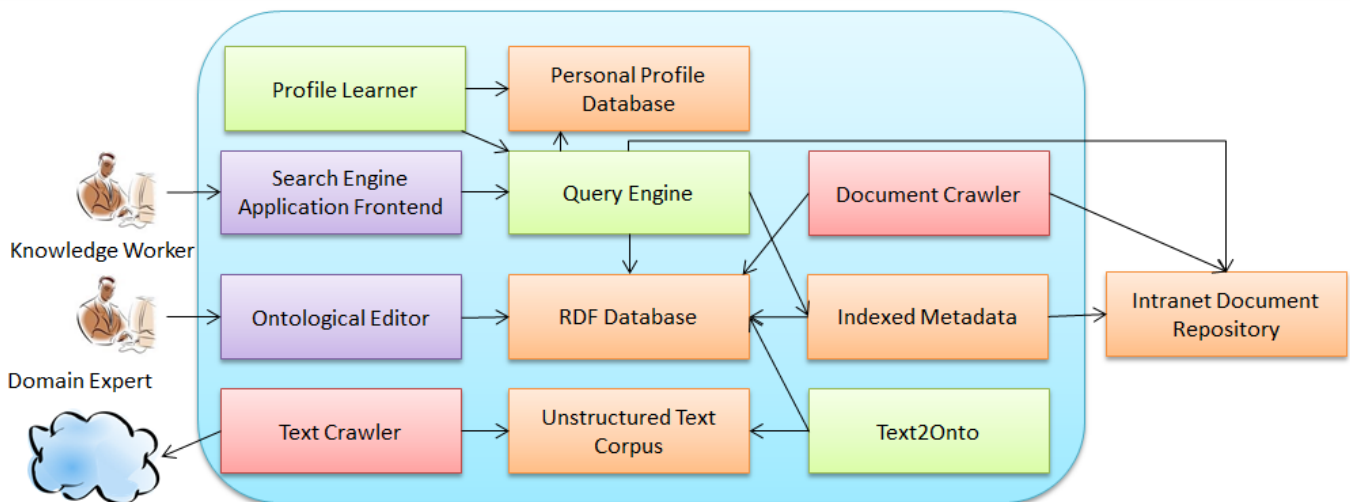

Fig. 1 Architecture

The domain ontology is used by the Document Crawler to index the existing document repository, containing the relevant company data and available on intranet. We chose to keep the metadata separate from the indexed documents, in a separate registry "Indexed Metadata" which contains maps between the concepts and the documents indexed with those concepts. The result of internal crawling has a feedback on the domain ontology, promoting concepts found in the corporate knowledge repository.

The real-time and most visible component of the system is the search engine itself. The engine uses a graphical interface to get the query term from the knowledge worker user and pass it to the query engine. The query engine uses the RDF database to expand the query with related concepts and then executes the query on the indexed metadata repository. The relevant results are presented to the user and when the user selects a result the respective document is retrieved from the intranet document repository.

Subsequent queries and chosen results from the same user are analyzed by the profile learner which discovers the user preferences and sub-domain of interest. The user profile is a map between concepts from the RDF database and ratings, promoting the concepts with greater selection rate. The ratings are used in the next queries for ranking the found results.

## IV. ONTOLOGY GENERATOR COMPONENT

For our prototype we have used Text2Onto as the main part of our ontology generator, have made some improvements to it and have benchmarked the results. We have chosen to compare the two types of ontologies within two distinct fields: insurance and software.

When automatically building ontologies from text, composing the corpus itself required manual input. This was performed by drawing the information from several sources relevant to the domain of the desired resulting ontology. In the case of our chosen fields, this involved theoretical descriptions, commercial presentations of products and services, as well as technical articles. The text originated from web pages, and PDF files found online. For our experiment we targeted a reasonable processing time (under one hour). In this setting we have found the maximum size of the corpus that could be processed by Text2Onto to be approximately 90000 characters (15000 words). Consequently we constructed the corpus to be as close as possible to the discovered size threshold.

### A. Method

The comparison also required for us to search for manually-built ontologies that outlined the selected fields with the granularity specific to domain ontologies. We have used Swoogle to search for existing ontologies. Our attempt at finding ontologies that matched our established criteria proved to be challenging, as the resources for ontologies describing areas of expertise were fewer than expected. The sources are therefore deemed to be somewhat unreliable and this reflects on the quality of the manually-constructed ontologies that were used in the comparison process.

Evaluating the similarity between the two ontologies is done through a "common concepts" percentage value. The "common concepts" percentage is computed using the following formula:

$$cc\ (\%) = (No.\ common\ concepts) / (Total\ no.\ concepts\ in\ both\ ontologies)$$

where $cc$ represents the "common concepts" percentage. This formula is based on the Jaccard similarity coefficient, which has the purpose to compare the similarities between sets.

The computation thus implies finding the number of concepts that are common between both ontologies. This is achieved by iterating through every concept in the automatic ontology and, for each concept, generating a list of all concepts synonymous to the selected concept. We search for any intersection between the set of concepts in the manual ontology and the concepts and resulting synonyms in the automatic ontology.

For our evaluation we have compared the manual with the

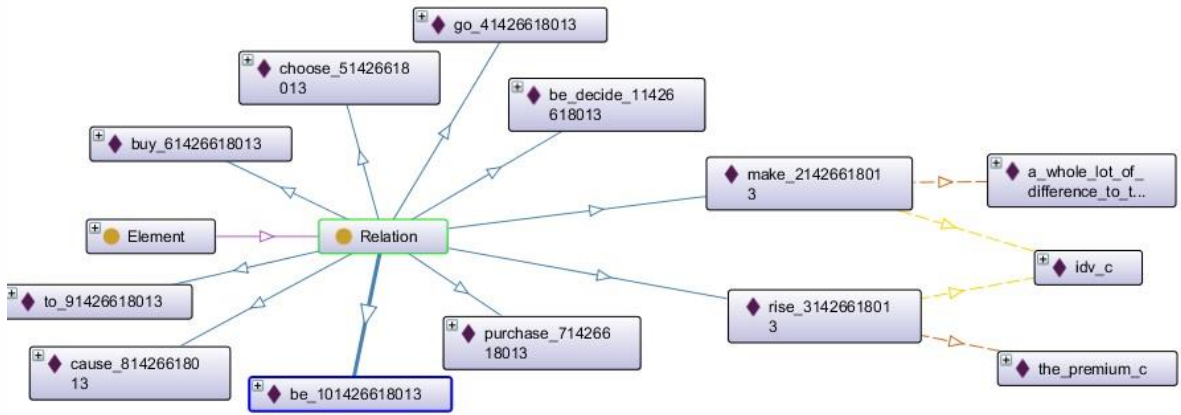

Fig. 2 Visualization of added relations

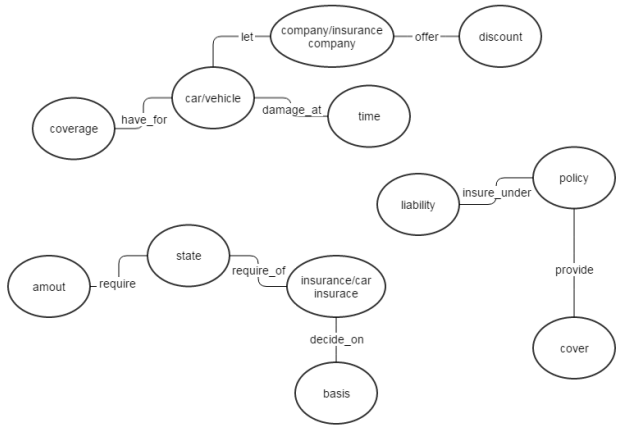

Fig. 3 Found concepts and relations in Insurance ontology

generated ontology in four cases:

- static threshold, in which we have set a lower bound for the probability of the concepts returned by Text2Onto and have kept that bound for all concepts.

- dynamic threshold, in which we have used a semi-automatic approach, by lowering or raising the lower bound in a set range for the concepts using human input.

- static threshold with similarity reduction, we have kept the lower bound fixed and have used our implemented similarity algorithm to trim the generated ontology.

- dynamic threshold with similarity reduction, we have used human input to make minor adjustments to the lower bound and have used our similarity algorithm to trim the generated ontology.

*B. Implementation*

One of the aspects improved from the existing implementation of the Text2Onto framework is the evaluation of concept similarity. We introduced an algorithm which analyzes similarity in terms of concepts that are synonyms to other another. If the concepts are synonymous, their similarity index is set to 1.0, otherwise their cosine similarity is computed. This was implemented by making use of JAWS, a Java library that enables users to manipulate data and perform operations from WordNet. Through this, we were able to compare two concepts at a time and decide whether or not they were synonymous. For the first concept, an array of its noun synsets would be generated. Each of these synsets would, in turn, have a list of word forms constructed. The final list of synonyms would be comprised of the total set of word forms of all the synsets of a concept. The algorithm would then verify whether the second concept belonged in that list of synonyms. The relation representing synonymous concepts is symmetric, so there is no need to further ensure that the second concept has the first concept as a synonym. In Table I, the concepts which were found similar using our algorithm are highlighted with bold. As seen, they are more relevant than the ones found using the default algorithm.

TABLE I. SIMILAR CONCEPTS

| Concept | Similar concept |
|---|---|
| **policy** | **insurance** |
| vehicle | person |
| vehicle | incident |
| person | incident |
| **person** | **someone** |
| excess | person |
| excess | incident |
| excess | vehicle |

In Text2Onto the extraction of relations is based on the part-of-speech (POS) tagging, when nouns are attributed with one of the tags NN, NNS, NNP, NNPS, depending on their singular or plural form. Verbs are tagged according to their tense using VB(verb, form), VBD(verb, past tense), VBG(verb, at gerund or present participle form), VBN(verb, past participle), VBP(verb, non third person, singular form, at present tense), VBZ(verb, third person, singular, at present tense). The tag JJ is used for adjectives, with IN is applied for prepositions or for a subordinating conjunction.

The software GATE (General Architecture for Text Engineering) is an open source software that allows designing and processing ontologies. The component that recognizes regular expressions in annotations is JAPE (Java Annotation Patterns Engine).

Originally, Text2Onto does not provide more than 3 rules for extracting relations, but starting from the given ones, we have extended them into new rules. For a rule to be more flexible such that it can extract several relations, the operator "*" can be use at the end of a declaration, such that the specific part of the phrase can be inexistent, or it can appear multiple times. A new rule is declared using the format "Rule:Rule1", followed by the declaration of the phrase parts. While declaring a verb, which is followed by another verb which may appear or not, like in "The paper is analyzed by providing…", the following format is used: ((({Token.category == VBD} | {Token.category == VBG} | {Token.category == VB} | {Token.category == VBN}):verb)*, where in the given example, the second verb is at the gerund form, resulting in the tag VBG.

There are cases when the vocabulary existent for the POS has to be extended by providing new words in order to be recognized while parsing the text and analyzing the tokens.

A relation is composed of a domain and range. For the phrase "If you raise the IDV, the premium rises.", the found relation by the program is rise. For the same subject, in the phrase "IDV can make a whole lot of difference to the motor insurance premium.", the extracted relation is make(idv, a whole lot of difference to the motor insurance premium).

The issue that appears in Text2Onto is that for multiple appearances of the same verb, just the first occurrence is saved, because it already appears in the hashmap as being stored. As mentioned in the paper [15], the domain and the range can be concatenated with the new found words by the use of a separator. The next step which follows ontology generation is having as many domains and ranges according to the number of the separators incremented plus one, because when one separator appears it splits the domain or range into the left-hand side and the right-hand side.

TABLE II. INSURANCE DOMAIN

| Case | Common concepts ( %) | No. generated concepts above threshold | No. concepts manual ontology | No. common concepts |
|---|---|---|---|---|
| Static threshold | 3.472 | 117 | 27 | 5 |
| Variable threshold | 7.352 | 41 | 27 | 5 |
| Static threshold with similarity reduction | 3.816 | 104 | 27 | 5 |
| Variable threshold with similarity reduction | 8.196 | 34 | 27 | 5 |

TABLE III. SOFTWARE DOMAIN

| Case | Common concepts ( %) | No. generated concepts above threshold | No. concepts manual ontology | No. common concepts |
|---|---|---|---|---|
| Static threshold | 5.150 | 423 | 276 | 36 |
| Variable threshold | 13.802 | 79 | 276 | 49 |
| Static threshold with similarity reduction | 5.723 | 353 | 276 | 36 |
| Variable threshold with similarity reduction | 14.939 | 52 | 276 | 49 |

The visualization of some of the few relations extracted has been done using Protégé, an open source ontology editor. In Fig. 2 the two relations mentioned before have the same domain and this is noticeable by the yellow color, the range being pointed with the orange color.

Another addition in the project was automatizing the process of finding similar concepts between the two types of ontologies. The concepts found by Text2Onto were returned as a list of POMEntity objects, which describe the label and relevance of the concept within the corpus. This list would further either be filtered with a static lowest bound threshold value on the concepts' relevance or would be written in a separate file for further processing. The latter entails manually setting the relevance of each concept, either to 1.0 or 0.0, in terms of its importance to the overall domain. We will further on refer to this operation as setting a variable threshold for the concepts. This aspect of the process was only semi-automatic. It is important to note that this conforms to the intended usage of Text2Onto, in which the user provides the system with feedback on whether or not the selected information is considered to be valid. Concepts from the manually-constructed ontologies were extracted from the OWL file using the SPARQL query below.

*select distinct ?class where{*

*{ ?class a rdfs:Class } union*

   *{ ?class rdfs:subClassOf|^rdfs:subClassOf [] }*

*}*

Once we had processed the input lists of Text2Onto concepts for each case, we made a comparison between the concepts in these lists and the concepts from the manually-generated ontology. Comparing the two ontologies implies finding which concepts can be found in both ontologies. For each concept from the automatically-generated ontology, a list of its synonyms is constructed. A concept from the automatically-built ontology is said to be found in both ontologies if the concept itself or any of its synonyms can be found in the manual ontology.

Implementing the automatic detection of common concepts also involved the use of WordNet and its synonym functionalities. When searching for common concepts between the two types of ontologies, we would iterate through every concept in the automatic ontology and retrieve its list of synonyms. This was achieved much like in the concept similarity algorithm, by generating all word forms of all synsets of a concept. Once the list of synonyms belonging to a concept is assembled, the algorithm saves any concepts that are at the intersection between the set of synonyms and the set of concepts from the manual ontology. The resulting set is appended to the overall set of common concepts. This process is repeated for each concept in the automatically-generated ontology. Using the overall common concepts set and the input sets from the two ontologies, we compute the common concept percentage according to the formula described in the previous section.

C. Results

For the insurance domain the results are presented in Table II and for the software domain the results are presented in Table III. A snippet of the insurance ontology is also presented in Fig. 3.

For the first ontology the used static threshold was 0.00142% and for the second the threshold was 0.00056%.

When analyzing the results we must reiterate that the benchmark ontologies were found on Swoogle, not manually designed from the same corpus data. Even so the relevance was acceptable and our improvements to the framework are visible, especially in the relations and similar concepts area which are used for query expansion. The generated domain ontology acts only as a starting point, which will be further refined when actually indexing the company documents.

## CONCLUSIONS

The first phase of a semantic search engine to be used in enterprise knowledge management for the purpose of finding relevant documents was presented. The focus of this paper was the architecture draft of the whole project and the results of experimenting with ontology learning frameworks for the ontology generator component.

We have chosen to use Text2Onto and have made different improvements to raise the relevancy of the generated ontology. For the next phase we will focus on the metadata annotation of the documents in the enterprise repository, with direct application to technical solutions used by small and medium enterprises.